\documentclass[sigconf]{acmart}

\usepackage[ruled]{algorithm2e}

\AtBeginDocument{%
  \providecommand\BibTeX{{%
    \normalfont B\kern-0.5em{\scshape i\kern-0.25em b}\kern-0.8em\TeX}}}

\acmSubmissionID{}

\begin{document}

\title{PAT: Pseudo-Adversarial Training For Detecting Adversarial Videos}

\author{Nupur Thakur}
\email{nsthaku1@asu.edu}
\affiliation{
  \institution{Arizona State University}
  \city{Tempe}
  \state{Arizona}
  \country{USA}
}

\author{Baoxin Li}
\email{baoxin.li@asu.edu}
\affiliation{
  \institution{Arizona State University}
  \city{Tempe}
  \state{Arizona}
  \country{USA}
}

\begin{abstract}
  Extensive research has demonstrated that deep neural networks (DNNs) are prone to adversarial attacks. Although various defense mechanisms have been proposed for image classification networks, fewer approaches exist for video-based models that are used in security-sensitive applications like surveillance. In this paper, we propose a novel yet simple algorithm called Pseudo-Adversarial Training (PAT), to detect the adversarial frames in a video without requiring knowledge of the attack. Our approach generates `transition frames' that capture critical deviation from the original frames and eliminate the components insignificant to the detection task. To avoid the necessity of knowing the attack model, we produce `pseudo perturbations' to train our detection network. Adversarial detection is then achieved through the use of the detected frames. Experimental results on UCF-101 and 20BN-Jester datasets show that PAT can detect the adversarial video frames and videos with a high detection rate. We also unveil the potential reasons for the effectiveness of the transition frames and pseudo perturbations through extensive experiments. 
\end{abstract}


\settopmatter{printacmref=false}
\setcopyright{none}
\renewcommand\footnotetextcopyrightpermission[1]{}
\pagestyle{plain}
\maketitle

\section{Introduction}

Deep neural networks (DNNs) have proven to be excellent learners for various tasks like image classification and video action recognition. Recent studies have also shown the vulnerability of DNNs to adversarial attacks \cite{goodfellow2014explaining, moosavi2016deepfool, dong2018boosting}. An example is when the input is altered subtly, leading to misclassification by the DNN. This has drawn a lot of attention as DNNs are being used for applications like surveillance \cite{sudhakaran2017learning}, autonomous vehicles \cite{zhang2016instance, ondruska2016deep}, facial recognition \cite{schroff2015facenet, liu2017sphereface}, etc., where resilience to adversarial attacks is of utmost importance. 

While the current literature documents extensive research related to adversarial learning in image-based applications, \cite{kurakin2016adversarial, xie2019improving, baluja2017adversarial, moosavi2016deepfool, carlini2017towards, su2019one}. 
the adversarial vulnerability of video-based DNNs remains a less explored area. This poses a pressing practical challenge, as DNNs are widely deployed in various video-analysis tasks \cite{wang2019learning, wang2019ranet, newell2016stacked, simonyan2014two, donahue2015long, feichtenhofer2016convolutional}. Although the video-based DNNs are more difficult to attack due to the additional temporal dimension, \cite{wei2019sparse} showcased the susceptibility of video action recognition models to adversarial attacks. 

The attacks on such models can be categorized into two types - (1) sparse attack where either minimum number of frames are perturbed or minimum amount of perturbations are added to each frame like \cite{wei2019sparse}, (2) dense attack where perturbations are added to all the frames like in \cite{li2018adversarial}.

There are plenty of defenses designed to defend image attacks \cite{papernot2016distillation, tramer2017ensemble, lu2017safetynet, song2017pixeldefend, meng2017magnet}. However, in general, they cannot be directly applied to videos as they do not take temporal information into account. Often, defenses for videos need to be computationally efficient, in order to handle the volume of data. The temporal redundancy present in videos can lead to a significant waste of computation if one simply applies an image-based approach on a frame-by-frame basis. To overcome these challenges, we suggest that the detection of adversarial videos will be a better alternative as compared to techniques involving intensive training, changes to network parameters or reconstructing the frames \cite{lo2020defending}. 

In this paper, we propose a novel defense strategy, Pseudo-Adversarial Training (PAT), for video action recognition networks which can detect the adversarial frames without any prior knowledge of the perturbations. It is called `pseudo' because the network is not trained on the actual perturbations. The detection network is trained on transition frames and pseudo perturbations to detect the perturbed frames. The transition frames, constructed from the neighboring frames, make the otherwise finely blended perturbations noticeable. 
The pseudo perturbations enable the network to learn about potential deviations from authentic frames, without the need to know specific attack models. 

\begin{figure*}[h]
    \centering
    \includegraphics[width=\linewidth]{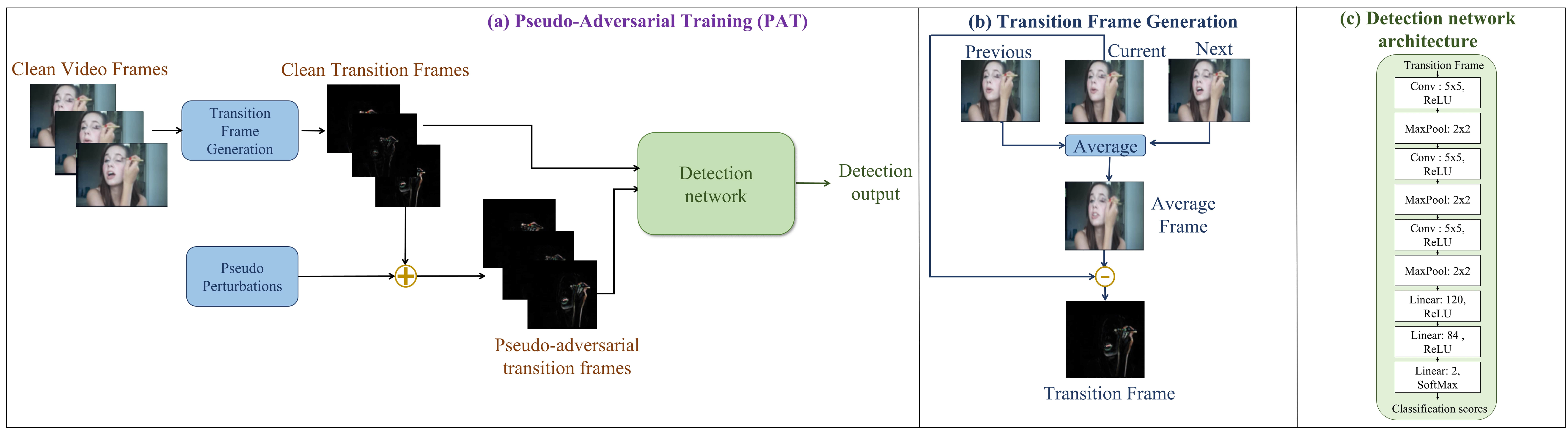}
    \caption{Overview of the Pseudo-Adversarial Training (PAT) approach. (a) shows the key components - transition frames generation, pseudo-perturbation generation and training the detection network using the clean transition frames and pseudo-adversarial transition frames. (b) shows transition frame generation. (c) shows the detection network architecture. }
    \label{architecture}
\end{figure*}

We summarize our contributions as follows: 
\begin{itemize}
    \item We propose a new technique, Pseudo-Adversarial Training (PAT) to detect the adversarial frames in a video. This strategy does not need prior knowledge of the attack scheme or the added perturbations, and can defend both the sparse as well as the dense attacks.   
    \item We define and use the transition frames so that the detection network can focus on perturbations that are more relevant to the detection of the adversarial frames rather than other elements in the frame.  
    \item We also propose to generate pseudo perturbations and use them to train our detection network. These perturbations are generated such that the network can learn to handle varying perturbations due to adversarials.   
\end{itemize}

\section{Related Work}

Ever since the adversarial attacks were discovered by \cite{szegedy2013intriguing}, they have been researched extensively \cite{zhao2017generating, goodfellow2014explaining, moosavi2017universal, papernot2016limitations} for various computer vision tasks. Early attacks like \cite{goodfellow2014explaining, carlini2017towards, moosavi2016deepfool} used the gradient information of the image models to generate the perturbations. As having the access to the network parameters is a strong assumption, the black-box attacks like \cite{papernot2017practical, liu2016delving} on the image models were designed which used the principle of transferability \cite{papernot2016transferability} to fool the network. To defend these attacks, one simple technique was to retrain the network with the adversarial images \cite{madry2017towards}. As the attacks grew stronger \cite{kos2018adversarial, dong2018boosting, zeng2019adversarial}, advanced defenses involving architectures like Generative Adversarial Networks (GANs) \cite{samangouei2018defense, song2017pixeldefend} were designed. 

The fact that the video action recognition models are vulnerable to adversarial attacks was first studied in \cite{wei2019sparse}. A video action recognition model aims to predict the label for an input video. Several deep learning models are used for this task like CNN+LSTM \cite{donahue2015long} etc. Such networks learn the spatial and temporal information from the video input. \cite{wei2019sparse} used $l_{2,1}$ optimization loss to generate the perturbations. They achieved temporal sparsity and a high fooling ratio using a temporal mask and propagation of the perturbations to the consecutive frames. 

\cite{rey2018targeted} makes use of a neural network to generate perturbations that are rich in detail and sparse. As these perturbations are highly specific, they are created for each video separately. \cite{li2018adversarial} is a white-box attack which makes GAN structure to generate `circular dual universal perturbations', with the discriminator being the target network. A post-processor is added between the generator and discriminator to perform circular shift on the perturbations. 

\cite{jiang2019black} is a black-box attack exploiting the pre-trained image models and black-box optimization techniques to minimize the queries to the target model and the search space. \cite{wei2020heuristic} is another black-box attack which takes a heuristic approach to key frames and regions to add the perturbations. 

The focus on the spatial information of the images in the image-based defenses makes them non-applicable to the video-based models. \cite{xiao2019advit} designed a defense for video-based models, where the optical flow is used to generate the current frame based on the previous. The video is then passed through the target network to check for temporal consistency.

\cite{jia2019identifying} also use the concept of temporal consistency to determine the perturbed frames in a video. They use a well-trained network to determine the labels of each frame and a frame is considered as adversarial if its label is different from its adjacent frames. \cite{lo2020defending} presents a defense method that replaces the batch normalization layers in the action recognition network with their module named MultiBN. They adversarially train this modified network to defend against adversarial attacks.  

PAT avoids the complex tasks of frame reconstruction and optical flow estimation by using instead the `transition frames', which are computed by a much simpler process. PAT neither introduces any hyperparameters for detection (like that in \cite{jia2019identifying}) nor requires retraining the target classifier (like that in \cite{lo2020defending}) and hence reduces the overhead of learning.

\begin{figure*}[h]
    \centering
    \includegraphics[width=\linewidth]{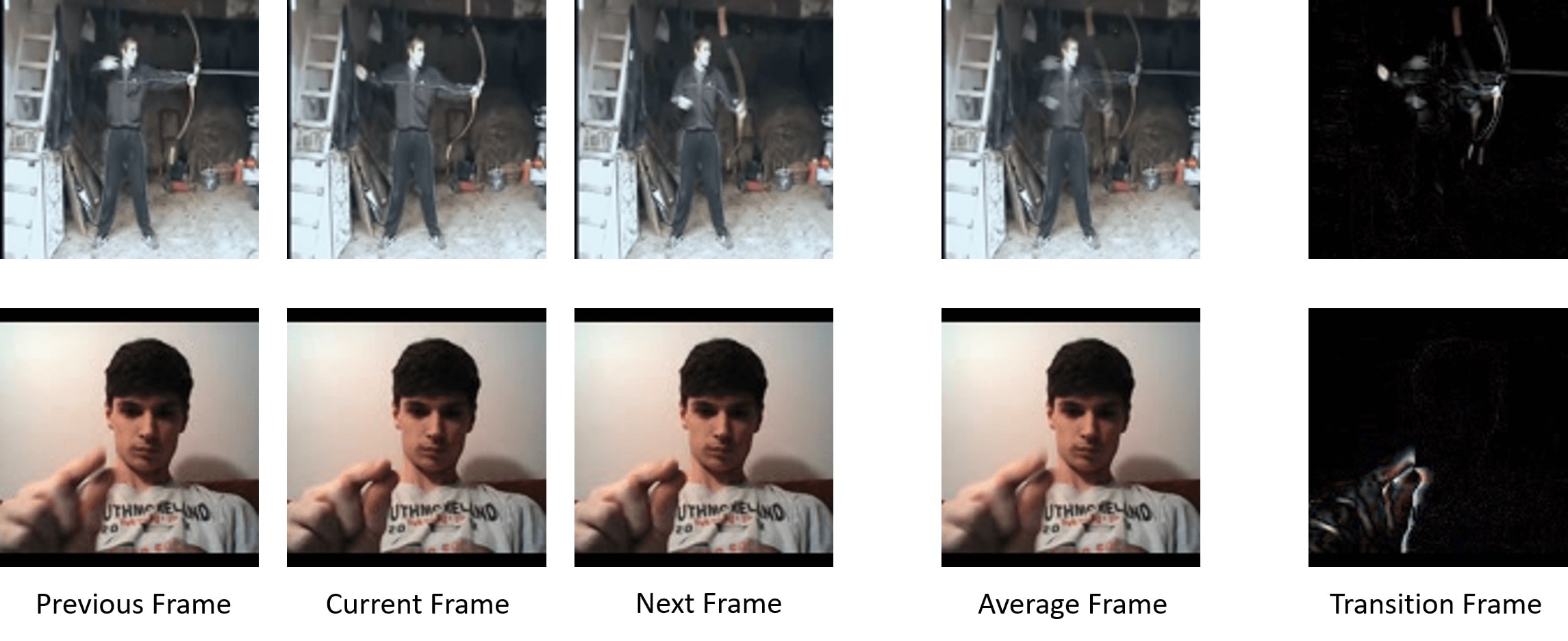}
    \caption{Sample original frames, the average frames and the transition frames. The first 3 columns show the original frames of videos. The $4^{th}$ column is the average of the previous and the next frames. The last column shows the transition frames for the current original frames. The transition frames eliminate most of the background and retain the object and the surrounding motion only. The frames in the $1^{st}$ row belong to UCF-101 dataset and the ones in the $2^{nd}$ row belong to 20-BN Jester dataset.}
    \label{diff_frame_pic}
\end{figure*}

\section{Methodology}

\subsection{Problem Definition}
Without loss of generality, we consider video action recognition models in this work but it can be easily extended to other video-related tasks. Let $X_{1}$, $X_{2}$,...., $X_{t}$, $X_{t+1}$ be the image frames of a video with $X_{t}$ being the target frame. Let $D$ be the classifier model and the prediction output of $X_{t}$ be $Y_{t}$ ($D(X_{t}) = Y_{t}$). An attacker aims to generate an adversarial frame $X_{t}^{*}$ by adding a small perturbation $\epsilon$ to the target frame $X_{t}$ such that $D(X_{t}^{*}) = Y_{t}^{*}$, where $Y_{t}^{*}$ is the adversarial target output. Our aim is to detect the adversarial frame $X_{t}^{*}$ without any knowledge of the attacking algorithm besides the two frames - $X_{t-1}$ and $X_{t+1}$. Also, there is no definite knowledge if $X_{t-1}$ and $X_{t+1}$ are clean or perturbed frames. 

\subsection{Pseudo-Adversarial Training (PAT)}

The Pseudo-Adversarial Training (PAT) strategy consists of three key components: transition frame generation, pseudo perturbation generation and training our detection network to detect the adversarial frames, which are elaborated below, with overall framework shown in Fig. \ref{architecture}.  

\subsubsection{Transition Frames Generation}

A video is a sequence of frames depicting a (dynamic) scene. The temporal component plays an important role in video-related tasks. Assuming a reasonable frame rate and small and continuous motion, an approximate reconstruction of the frame is possible using the nearby frames. PAT leverages these facts to compute `transition frames', which are used to capture the underlying dynamics across the frames while ignoring static portions of the frames. 
Consider three consecutive frames $X_{t-1}$, $X_{t}$ and $X_{t+1}$. We define `motion' $M_{1}$ between $X_{t-1}$ and $X_{t}$ as $M_{1} = X_{t} - X_{t-1}$. The term `motion' is used in this narrow sense throughout this paper, and it is supposed to capture underlying dynamics. Similarly, the motion $M_{2}$ between $X_{t}$ and $X_{t+1}$ is $M_{2} = X_{t+1} - X_{t}$. Now, the motion between $M_{1}$ and $M_{2}$ is given by - 

\begin{equation}
    M_{2} - M_{1} = X_{t+1} - X_{t} - X_{t} + X_{t-1}
    \label{m21}
\end{equation} 

Eq. \ref{m21} reduces to the transition frame equation (Eq. \ref{diff_frame}). These equations show how the transition frame is able to capture the motion from the three frames used to create it. It is generated using two simple operations only - average and difference, making it computationally inexpensive. 

\begin{equation}
    X_{t}^{tr} = \left(\frac{X_{t-1} + X_{t+1}}{2}\right) - X_{t}
    \label{diff_frame}
\end{equation}

There are two special cases - the first and the last frame of the video. The first frame does not have the previous frame and the last frame does not have the next frame. 
Thus, we replace the average frame for first and last frames by the second frame and the second last frame in the video respectively.

Fig. \ref{diff_frame_pic} shows the original frames, their average and the transition frames for the current frames. It is clear from the transition frame that it gets rid of most of the static background. It only contains the main object and the motion around it. Elimination of the passive elements does not hamper the detection process as they are not relevant to our detection task. As a result, the perturbations become visible prominently in the transition frames.

\subsubsection{Pseudo Perturbations Generation}

For training the detection network, we need clean and adversarial samples. As there is no prior knowledge of the perturbations being added or the attack algorithm, we propose a way to generate on-the-go pseudo perturbations. These perturbations are not actual perturbations as they are not generated using an attack strategy. But, when added to the transition frames, they are enough to let the network learn about the actual perturbations in the video frames. 

To generate the pseudo perturbations, we use a varying magnitude of Gaussian noise by changing the standard deviation $\sigma$ of the Gaussian distribution. We generate Gaussian noise mask $X_{n} \sim \mathcal{N}(0,\,\sigma)\,$ where $\sigma \sim \mathcal{U}(0.0001,\,0.05)\,$. It is of the same shape as that of the transition frame (Eq. \ref{pseudo_perturbations} where $X_{p_{adv}}$ represents the pseudo-adversarial transition frame). We choose this particular range for the value of $\sigma$ because it covers a wide variety of magnitude. For the values below 0.0001, the noise does not make any significant impact on the image. The values above 0.05 completely disrupt the transition frame and will turn it into noise only. For every transition frame, a different value of $\sigma$ is picked randomly. This varying noise mask added to the transition frames is essential to train the detection network such that it can identify a variety of perturbations.

\begin{equation}
    X_{p_{adv}}^{tr} = X_{t}^{tr} + X_{n} 
    \label{pseudo_perturbations}
\end{equation}

\subsubsection{Training the detection network}

We use a convolutional neural network (architecture shown in Fig.\ref{architecture} (c)) for binary classification as the detection network. The transition frames for the clean class are obtained from the original videos. For adversarial class, the transition frames are obtained by adding the pseudo perturbations to the clean transition frames. Our approach is summarized in Algorithm \ref{model_algorithm} \footnote{Please email nsthaku1@asu.edu for the code of this approach.}. 

\section{Experiments and Results}

In this section, we start with the discussion of the experimental settings and the metrics used for evaluation. We also discuss the two attack baselines used to evaluate our technique. Next, we present the results of PAT on two datasets - UCF-101 \cite{soomro2012ucf101} and Jester \cite{materzynska2019jester} dataset. 

\subsection{Experimental Settings}
\subsubsection{Datasets and Target Networks} 
We chose UCF-101 \cite{soomro2012ucf101} and Jester \cite{materzynska2019jester} dataset for our experiments to showcase that our approach works for both coarse-grained action recognition and fine-grained action recognition data. UCF-101 dataset contains 101 human-action classes (coarse-grained) like playing cello, applying lipstick, hammering etc. It contains variations in terms of illumination conditions, background, camera etc. The dataset comes with 3 different train-test splits. We use split 1 consisting of 9,537 training videos and 3,783 testing videos for all our experiments. 

The 20BN-Jester dataset contains hand-gesture videos where people perform a particular hand-gesture in front of a webcam or camera. It has 27 fine-grained categories of hand gestures like zooming in with the full hand, zooming in with two fingers, zooming out with full hand etc. There are 118,562 training videos, 14,787 validation videos and 14,743 testing videos. As the labels for testset are not available, we provide the results on the validation set.  

\begin{algorithm}
\SetAlgoLined
\SetKwData{Left}{left}\SetKwData{This}{this}\SetKwData{Up}{up}
\SetKwFunction{Union}{Union}\SetKwFunction{FindCompress}{FindCompress}
\SetKwInOut{Input}{Input}\SetKwInOut{Output}{Output}
\Input{Training Videos $\bf{X}$, Labels $y_c$ and $y_{adv}$, Test Videos $\bf{X_{test}}$, Detection Network $D$}
\Output{Predicted Labels $\hat{y}$}
\For{each epoch $e = 1,2,...$}{
\While{Training}{
\For{each video frame $X_{t} = 1,2,...$}{
$X_{t}^{tr} = \left(\frac{X_{t-1} + X_{t+1}}{2}\right) - X_{t}$\;
Generate $\sigma \sim \mathcal{U}(0.0001,\,0.05)\,$\;
Generate $X_{n}\sim \mathcal{N}(0,\,\sigma)\,$\;
$X_{p_{adv}}^{tr} = X_{t}^{tr} + X_{n}$\;
Train $D$ using cross-entropy loss; }}
\While{Testing}{
\For{each video frame $X_{test_t} = 1,2,...$}{
$X_{test_t}^{tr} = \left(\frac{X_{test_{t-1}} + X_{test_{t+1}}}{2}\right) - X_{test_t}$\;
$\hat{y} = D(X_{test_t}^{tr})$; }}
}
\caption{PAT: Pseudo-Adversarial Training}
\label{model_algorithm}
\end{algorithm}

The target network for UCF-101 dataset is CNN+LSTM classifier. The pretrained ResNet152 is followed by a layer of LSTM, plus two fully-connected layers and the final classification using softmax. We achieve a test accuracy of 91.09\%. The target network for Jester dataset is a C3D classifier. We fine-tune on the model of depth 18 pretrained on Kinetics dataset available on GitHub repository \cite{hara3dcnns}. Validation accuracy is 90.33\%. The input frame resolution is $112\times112$ for both the datasets. The sequence length is 40 frames and 16 frames for UCF-101 and Jester dataset respectively.  

\subsubsection{Attack Baselines}

We consider two attacks to evaluate our approach. Sparse Adversarial Perturbations \cite{wei2019sparse} perturbs only a small percentage of frames from the entire video using an $l_{2,1}$ optimization loss and temporal mask. As the perturbations are sparse, we refer to it as 'sparse attack' for the rest of the paper. For our experiments, we perturb pre-determined 22.5\% (9 out of 40 frames) and 20\% (4 out of 16 frames) of the total frames for UCF-101 and Jester dataset respectively. 

The other attack \cite{li2018adversarial} perturbs all the frames in a video using a generative model.  
As all the frames are perturbed, we refer to it as 'dense attack'. With these two attacks, we show that our method can detect adversarial frames containing a variety of perturbations. 

\subsubsection{Evaluation metrics}
Two evaluation metrics are used in our experiments: 1) Frame Detection Rate - 
\begin{equation}
    FDR = \frac{\sum_{i=0}^{N}D(X_{i}) = Y_{i}}{N}
\end{equation}
where $X_{i}$ is the input frame, $Y_{i} \in \{0,1\}$ is the ground truth for frame detection, D is Detection network and N is the total number of frames; 2) Video Detection Rate - 
\begin{equation}
    VDR = \frac{\sum_{i=0}^{M} \left(\prod_{j=0}^{N} D(X_{ij})\right) = Y_{i}}{M}
\end{equation}
where $X_{ij}$ is the $j^{th}$ frame in $i^{th}$ video, $Y_{i} \in \{0,1\}$ is the ground truth for video detection, D is Detection network, N is the number of frames and M is the number of videos. 

We also calculate Area under Receiver Operating Characteristic (ROC), shortly known as AUC on UCF-101 dataset for comparison purposes. AUC represents the degree to which the model can distinguish between two classes and ROC is a probability curve. Higher the Area under Curve (AUC), better is the capability of the model to distinguish between the two classes.

\subsection{Adversarial Detection Results}

Table \ref{detection_results} summarizes our results for adversarial detection on both the datasets. From the FDR column, it is clear that PAT detects adversarial frames from both types of attacks with high rate. This demonstrates that PAT can detect different types and magnitude of perturbations without having any prior knowledge about them. Also, PAT works well for both the datasets showing that it can handle coarse-grained and fine-grained classification data. This makes it a very promising approach for detecting the adversarial frames. Based on the detected adversarial frames, the adversarial video input can be detected. The frame detection rate obtained by PAT is enough to detect most of the adversarial videos with ease. 

\begin{table}[h]
\caption{Adversarial Frame detection rate (FDR) and video detection rate (VDR) for PAT on UCF-101 and 20BN-Jester Dataset for different attacks. }
\label{detection_results}
\centering
\begin{tabular}{|c|c|c|c|}
    \toprule
    \bf{Attack Algorithm} & \bf{FDR} & \bf{VDR}\\
    \toprule
    \multicolumn{3}{|c|}{\bf{UCF-101}}\\
    \midrule
    Sparse Attack  &83.62\%  &92.86\%\\
    \midrule
    Dense Attack  &83.46\%  &88.25\% \\   
    \midrule
    \multicolumn{3}{|c|}{\bf{20BN-Jester}}\\
    \midrule
    Sparse Attack  &75.9\%  &80.7\%\\
    \bottomrule
\end{tabular}
\end{table}

The sparse attack showed that even if one frame is perturbed, a success rate of 60\% is achieved. Therefore, even if one adversarial frame is detected in the entire video sequence, the video can be categorized as an adversarial one. However, to accommodate the scenario of having false positives, we use a threshold of 3 adversarial frames i.e we consider a video to be adversarial when atleast 3 frames are classified as adversarial by the detection network. See Table \ref{detection_results} for the adversarial VDR results for PAT. 

We tried different values of the threshold for adversarial frames to determine an adversarial video. We found empirically that a value of 3 for the threshold maintained a balance between false positives and false negatives. Higher values of threshold led to higher number of adversarial videos being misclassified as clean which can pose a threat to the network. On the other hand, lower values of threshold led to higher number of clean videos being classified as adversarial, which is not desirable too.  

\begin{figure}[h]
    \centering
    \includegraphics[scale=0.58]{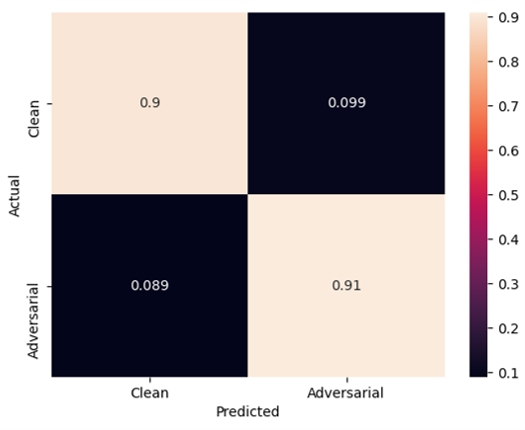}
    \caption{Confusion matrix of PAT for the adversarial video detection when the adversarial videos are generated by both sparse and dense attack. }
    \label{cm}
\end{figure}

Fig. \ref{cm} shows the confusion matrix for detecting adversarial video using PAT on UCF-101 dataset. The adversarial videos contain a mix of videos generated by sparse and dense attack. The high true positives and true negatives along with low misclassification of videos for both the classes indicate the ability of PAT to detect adversarial videos without any prior knowledge of perturbations.  

\begin{table}[h]
\caption{Comparison of PAT with other defenses (AUC) on UCF-101 dataset. The first column determines the type of defense mechanism used for defending the network against the adversarial videos. The last 3 columns determine the attack used to generate the test adversarial videos. The second last row is when the detection network is trained on set of clean and adversarial videos generated from sparse and dense attack, where the last row shows the results for the proposed method. }
\label{auc_results}
\centering
\begin{tabular}{|c|c|c|c|c|}
    \toprule
    \bf{Defense} & \bf{Sparse} & \bf{Dense}   & \bf{Sparse+Dense}\\
    \toprule
    Temporal+Spatial\cite{jia2019identifying}  &-  &-  &77\%\\
    \midrule
    AdvIT\cite{xiao2019advit}  &97\%  &-  &-\\
    \midrule
    PAT (Adversarial data)   &93.4\%  &99.8\%  &99.8\%\\
    \midrule
    PAT  &\textbf{97.6\%}  &\textbf{94.1\%}   &\textbf{94.2\%}\\
    \bottomrule
\end{tabular}
\end{table}

\begin{figure}[h]
    \centering
    \includegraphics[scale=0.47]{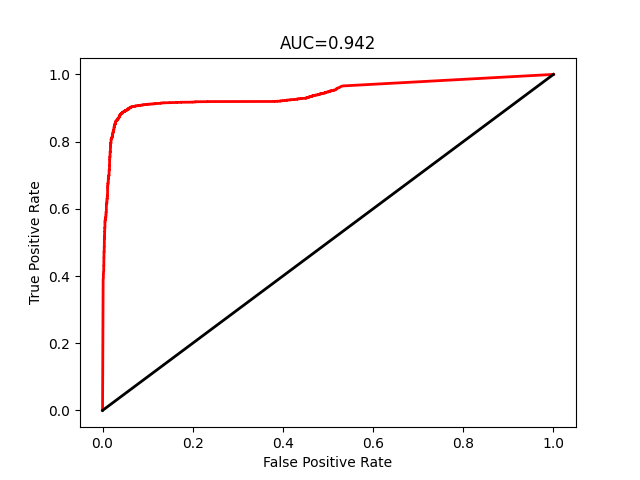}
    \caption{ROC Curve under different thresholds for performance of proposed method, PAT on clean and adversarial videos.}
    \label{roc}
\end{figure}

We also evaluate PAT using Area under Receiver Operating Characteristic curve (AUC) for comparing with other baselines. Table \ref{auc_results} shows the AUC results and the ROC curve for PAT is displayed in Fig. \ref{roc}. This curve indicates that our approach has a high capability in differentiating between clean and adversarial videos. PAT achieves AUC of 94.2\% on clean and adversarial (generated using both sparse and dense attack) videos.

In Table \ref{auc_results}, the first row determines the performance of PAT when it is trained using the clean and adversarial videos. It is not surprising that the performance on both the attacks is high in this case as the network is aware of the perturbations. PAT can achieve almost as good performance (94.2\% AUC which is just $\sim5\%$ lower) as the first case in Table \ref{auc_results}, without actually having the knowledge of the real perturbations. Our method also outperforms \cite{jia2019identifying} and \cite{xiao2019advit} with an improvement of approximately 17\% and 0.6\% respectively.       
\section{Ablation Study}

In this section, we showcase the importance of the two major components of the PAT algorithm - the transition frames and the varying value of standard deviation to generate the pseudo perturbations. We show how crucial these components are towards making our approach work well while being computationally inexpensive, effective, simple to implement at the same time using UCF-101 dataset. 

\subsection{Input Frame Analysis}

\begin{figure}[h]
    \centering
    \includegraphics[width=\linewidth]{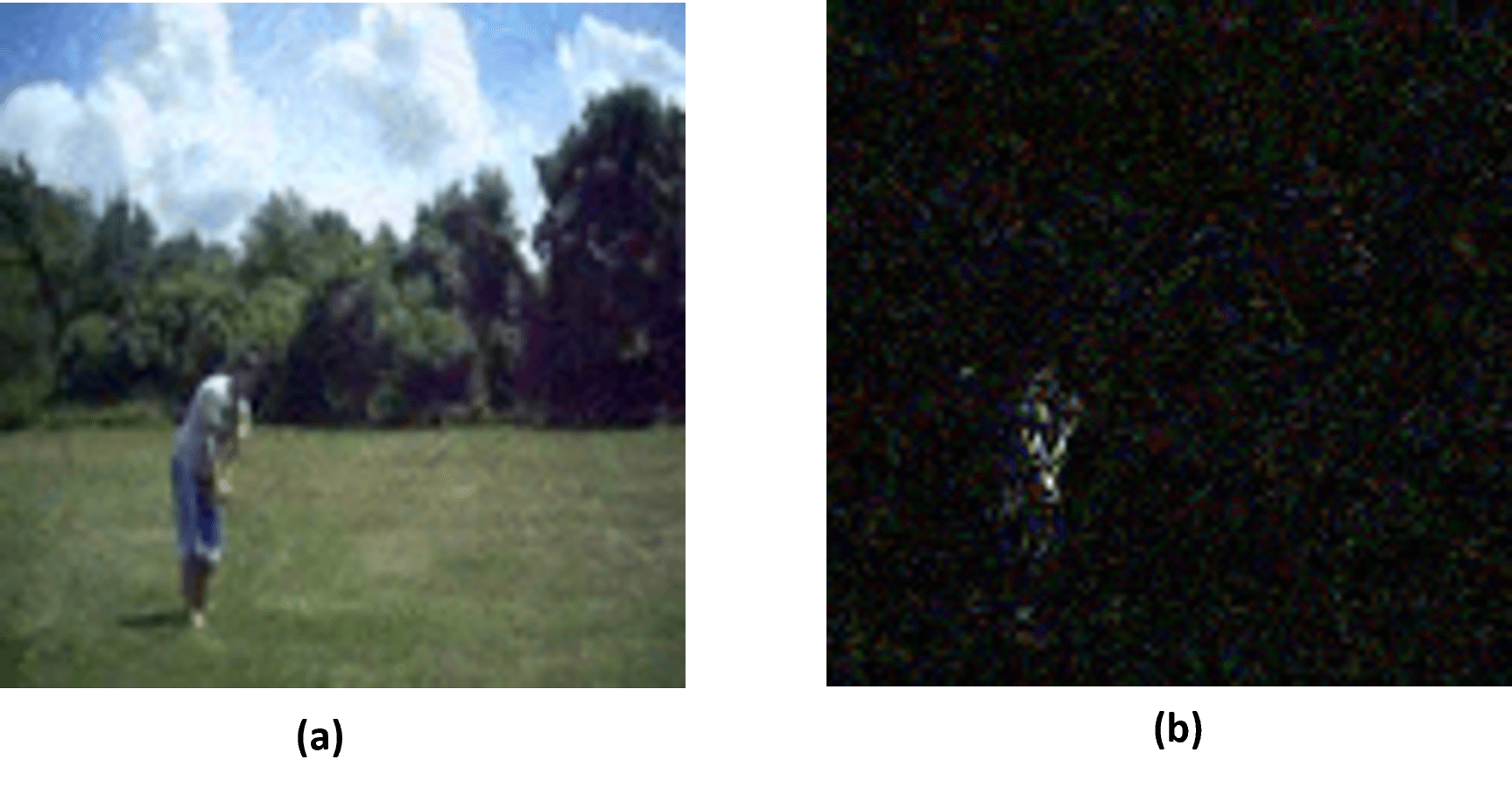}
    \caption{(a) is an adversarial video frame from a video belonging to UCF-101 dataset \& (b) is the corresponding transition frame. Perturbations are clearly visible in the transition frame while they are not noticeable in (a).}
    \label{adv_transition_frame}
\end{figure}

Factors like color, texture and background add complexity to a frame. The attackers take advantage of such components to blend in the added perturbations. This is where the transition frames play a major role in bringing the perturbations into the light. The transition frames, calculated using Eq. \ref{diff_frame}, eliminate the passive components in a frame and focus on the objects and their motion only. As a result, the perturbations have no way to blend in and therefore are clearly visible.   

Fig. \ref{adv_transition_frame} shows an adversarial frame from a video belonging to UCF-101 dataset and its corresponding transition frame. The perturbations in the left image do not stand out while being prominently visible in the transition frame (on the right). As our main goal is to detect the adversarial frame, many components like the background become insignificant. As such insignificant components are removed in the transition frame, the perturbations can be viewed as well as processed by a network easily. This also helps in keeping the detection network architecture small reducing the training and inference time. For all the above reasons, transition frames are an essential part of keeping the PAT algorithm simple and accurate. 

The $1^{st}$ row of Table \ref{analysis_results} shows the detection rate when original frames are used to train the detection network instead of the transition frames. The detection rate is almost equal to a random guess for both the attacks. This is because the original frames have a lot of information irrelevant for the detection task which acts as a perfect disguise for the perturbations. On the contrary, the transition frame keeps only relevant elements and has a higher adversarial frame detection rate.

\subsection{Pseudo Perturbations Analysis}

In PAT, the pseudo adversarial transition frames are generated using the Eq. \ref{pseudo_perturbations}. The standard deviation $\sigma$ is drawn from a uniform distribution and is different for every frame for each training epoch. This is crucial because the PAT can handle adversarial input with different perturbations.  

\begin{table}[h]
\caption{Adversarial FDR showing the importance of key components of PAT using UCF-101. Sparse and Dense are the two attack baselines. PAT with transition frames and varying $\sigma$ outperform all other cases. *$\sigma$ varies between 0.0001-0.05.}
\label{analysis_results}
\centering
\begin{tabular}{|c|c|c|c|}
    \toprule
     {\bf Frame Type} &{\bf $\sigma$} &\bf{Sparse} &\bf{Dense}\\
    \toprule
    Original &Varying* &52.31\% &51.12\%\\
    \midrule
    Transition &0.0001 &63.98\%  &49.6\% \\
    \midrule
    Transition &0.01 &80.12\% &64.65\% \\
    \midrule
    Transition &Varying* &\bf{83.62\%} &\bf{83.46\%} \\
    \bottomrule
\end{tabular}
\end{table}

The last 3 rows of Table \ref{analysis_results} show the effect of different values of $\sigma$ on the adversarial FDR with transition frames as input to the detection network. The adversarial FDR when the value of $\sigma$ is fixed throughout the training process is lower as compared to the one when the value of $\sigma$ varies for every frame and epoch during training since the fixed values do not cover the variations in the actual perturbations. With varying standard deviation, the network learns from a different version of the same transition frame during every epoch. Also, we observed that in some cases of fixed $\sigma$, the network overfits at some point of time.
For fixed $\sigma$ value of 0.01, the performance on the sparse attack is close to the best case FDR but the performance on the dense attack is poor. Thus, to achieve good performance on both the attacks, the transition frames and the varying standard deviation of Gaussian noise are both essential. 

\subsection{Run-time Analysis}
To showcase that PAT is computationally inexpensive, we empirically measure the running time for the our detection process using Nvidia Titan XP GPU. We use a mix of both clean and adversarial (sparse as well as dense attack) videos to determine the average detection time for PAT. Our approach takes 0.01 seconds on an average to determine whether a video is adversarial, which is extremely low overhead to the existing action recognition systems. 

\subsection{Conclusion}
We proposed a novel approach, PAT to detect the adversarial frames in a video efficiently and keep the video-based networks secure from different types of attacks. We achieve good detection rate without having any access to the attack or the perturbations, which is generally the case in real-world applications. Our experiments on UCF-101 and Jester datasets demonstrate that the approach is highly accurate in detecting the adversarial input produced by different attacks. We also show the detection of adversarial video based on the PAT detected frames. Furthermore, we demonstrated the importance of transition frames and the varying Gaussian noise to generate pseudo perturbations in achieving a good frame detection rate.

\bibliographystyle{ACM-Reference-Format}
\bibliography{sample-base}


\begin{thebibliography}{46}


\ifx \showCODEN    \undefined \def \showCODEN     #1{\unskip}     \fi
\ifx \showDOI      \undefined \def \showDOI       #1{#1}\fi
\ifx \showISBNx    \undefined \def \showISBNx     #1{\unskip}     \fi
\ifx \showISBNxiii \undefined \def \showISBNxiii  #1{\unskip}     \fi
\ifx \showISSN     \undefined \def \showISSN      #1{\unskip}     \fi
\ifx \showLCCN     \undefined \def \showLCCN      #1{\unskip}     \fi
\ifx \shownote     \undefined \def \shownote      #1{#1}          \fi
\ifx \showarticletitle \undefined \def \showarticletitle #1{#1}   \fi
\ifx \showURL      \undefined \def \showURL       {\relax}        \fi
\providecommand\bibfield[2]{#2}
\providecommand\bibinfo[2]{#2}
\providecommand\natexlab[1]{#1}
\providecommand\showeprint[2][]{arXiv:#2}

\bibitem[\protect\citeauthoryear{Baluja and Fischer}{Baluja and
  Fischer}{2017}]%
        {baluja2017adversarial}
\bibfield{author}{\bibinfo{person}{Shumeet Baluja} {and} \bibinfo{person}{Ian
  Fischer}.} \bibinfo{year}{2017}\natexlab{}.
\newblock \showarticletitle{Adversarial transformation networks: Learning to
  generate adversarial examples}.
\newblock \bibinfo{journal}{\emph{arXiv preprint arXiv:1703.09387}}
  (\bibinfo{year}{2017}).
\newblock


\bibitem[\protect\citeauthoryear{Carlini and Wagner}{Carlini and
  Wagner}{2017}]%
        {carlini2017towards}
\bibfield{author}{\bibinfo{person}{Nicholas Carlini} {and}
  \bibinfo{person}{David Wagner}.} \bibinfo{year}{2017}\natexlab{}.
\newblock \showarticletitle{Towards evaluating the robustness of neural
  networks}. In \bibinfo{booktitle}{\emph{2017 ieee symposium on security and
  privacy (sp)}}. IEEE, \bibinfo{pages}{39--57}.
\newblock


\bibitem[\protect\citeauthoryear{Donahue, Anne~Hendricks, Guadarrama, Rohrbach,
  Venugopalan, Saenko, and Darrell}{Donahue et~al\mbox{.}}{2015}]%
        {donahue2015long}
\bibfield{author}{\bibinfo{person}{Jeffrey Donahue}, \bibinfo{person}{Lisa
  Anne~Hendricks}, \bibinfo{person}{Sergio Guadarrama}, \bibinfo{person}{Marcus
  Rohrbach}, \bibinfo{person}{Subhashini Venugopalan}, \bibinfo{person}{Kate
  Saenko}, {and} \bibinfo{person}{Trevor Darrell}.}
  \bibinfo{year}{2015}\natexlab{}.
\newblock \showarticletitle{Long-term recurrent convolutional networks for
  visual recognition and description}. In \bibinfo{booktitle}{\emph{Proc. IEEE
  Conf. Computer Vision and Pattern Recognition}}. \bibinfo{pages}{2625--2634}.
\newblock


\bibitem[\protect\citeauthoryear{Dong, Liao, Pang, Su, Zhu, Hu, and Li}{Dong
  et~al\mbox{.}}{2018}]%
        {dong2018boosting}
\bibfield{author}{\bibinfo{person}{Yinpeng Dong}, \bibinfo{person}{Fangzhou
  Liao}, \bibinfo{person}{Tianyu Pang}, \bibinfo{person}{Hang Su},
  \bibinfo{person}{Jun Zhu}, \bibinfo{person}{Xiaolin Hu}, {and}
  \bibinfo{person}{Jianguo Li}.} \bibinfo{year}{2018}\natexlab{}.
\newblock \showarticletitle{Boosting adversarial attacks with momentum}. In
  \bibinfo{booktitle}{\emph{Proceedings of the IEEE conference on computer
  vision and pattern recognition}}. \bibinfo{pages}{9185--9193}.
\newblock


\bibitem[\protect\citeauthoryear{Feichtenhofer, Pinz, and
  Zisserman}{Feichtenhofer et~al\mbox{.}}{2016}]%
        {feichtenhofer2016convolutional}
\bibfield{author}{\bibinfo{person}{Christoph Feichtenhofer},
  \bibinfo{person}{Axel Pinz}, {and} \bibinfo{person}{Andrew Zisserman}.}
  \bibinfo{year}{2016}\natexlab{}.
\newblock \showarticletitle{Convolutional two-stream network fusion for video
  action recognition}. In \bibinfo{booktitle}{\emph{Proceedings of the IEEE
  conference on computer vision and pattern recognition}}.
  \bibinfo{pages}{1933--1941}.
\newblock


\bibitem[\protect\citeauthoryear{Goodfellow, Shlens, and Szegedy}{Goodfellow
  et~al\mbox{.}}{2014}]%
        {goodfellow2014explaining}
\bibfield{author}{\bibinfo{person}{Ian~J Goodfellow}, \bibinfo{person}{Jonathon
  Shlens}, {and} \bibinfo{person}{Christian Szegedy}.}
  \bibinfo{year}{2014}\natexlab{}.
\newblock \showarticletitle{Explaining and harnessing adversarial examples}.
\newblock \bibinfo{journal}{\emph{arXiv preprint arXiv:1412.6572}}
  (\bibinfo{year}{2014}).
\newblock


\bibitem[\protect\citeauthoryear{Hara, Kataoka, and Satoh}{Hara
  et~al\mbox{.}}{2018}]%
        {hara3dcnns}
\bibfield{author}{\bibinfo{person}{Kensho Hara}, \bibinfo{person}{Hirokatsu
  Kataoka}, {and} \bibinfo{person}{Yutaka Satoh}.}
  \bibinfo{year}{2018}\natexlab{}.
\newblock \showarticletitle{Can Spatiotemporal 3D CNNs Retrace the History of
  2D CNNs and ImageNet?}. In \bibinfo{booktitle}{\emph{Proc. IEEE Conf. on
  Computer Vision and Pattern Recognition}}. \bibinfo{pages}{6546--6555}.
\newblock


\bibitem[\protect\citeauthoryear{Jia, Wei, and Cao}{Jia et~al\mbox{.}}{2019}]%
        {jia2019identifying}
\bibfield{author}{\bibinfo{person}{Xiaojun Jia}, \bibinfo{person}{Xingxing
  Wei}, {and} \bibinfo{person}{Xiaochun Cao}.} \bibinfo{year}{2019}\natexlab{}.
\newblock \showarticletitle{Identifying and Resisting Adversarial Videos Using
  Temporal Consistency}.
\newblock \bibinfo{journal}{\emph{arXiv preprint arXiv:1909.04837}}
  (\bibinfo{year}{2019}).
\newblock


\bibitem[\protect\citeauthoryear{Jiang, Ma, Chen, Bailey, and Jiang}{Jiang
  et~al\mbox{.}}{2019}]%
        {jiang2019black}
\bibfield{author}{\bibinfo{person}{Linxi Jiang}, \bibinfo{person}{Xingjun Ma},
  \bibinfo{person}{Shaoxiang Chen}, \bibinfo{person}{James Bailey}, {and}
  \bibinfo{person}{Yu-Gang Jiang}.} \bibinfo{year}{2019}\natexlab{}.
\newblock \showarticletitle{Black-box adversarial attacks on video recognition
  models}. In \bibinfo{booktitle}{\emph{Proc. ACM Intl. Conf. on Multimedia}}.
  \bibinfo{pages}{864--872}.
\newblock


\bibitem[\protect\citeauthoryear{Kos, Fischer, and Song}{Kos
  et~al\mbox{.}}{2018}]%
        {kos2018adversarial}
\bibfield{author}{\bibinfo{person}{Jernej Kos}, \bibinfo{person}{Ian Fischer},
  {and} \bibinfo{person}{Dawn Song}.} \bibinfo{year}{2018}\natexlab{}.
\newblock \showarticletitle{Adversarial examples for generative models}. In
  \bibinfo{booktitle}{\emph{2018 ieee security and privacy workshops (spw)}}.
  IEEE, \bibinfo{pages}{36--42}.
\newblock


\bibitem[\protect\citeauthoryear{Kurakin, Goodfellow, and Bengio}{Kurakin
  et~al\mbox{.}}{2016}]%
        {kurakin2016adversarial}
\bibfield{author}{\bibinfo{person}{Alexey Kurakin}, \bibinfo{person}{Ian
  Goodfellow}, {and} \bibinfo{person}{Samy Bengio}.}
  \bibinfo{year}{2016}\natexlab{}.
\newblock \showarticletitle{Adversarial examples in the physical world}.
\newblock \bibinfo{journal}{\emph{arXiv preprint arXiv:1607.02533}}
  (\bibinfo{year}{2016}).
\newblock


\bibitem[\protect\citeauthoryear{Li, Neupane, Paul, Song, Krishnamurthy,
  Chowdhury, and Swami}{Li et~al\mbox{.}}{2018}]%
        {li2018adversarial}
\bibfield{author}{\bibinfo{person}{Shasha Li}, \bibinfo{person}{Ajaya Neupane},
  \bibinfo{person}{Sujoy Paul}, \bibinfo{person}{Chengyu Song},
  \bibinfo{person}{Srikanth~V Krishnamurthy}, \bibinfo{person}{Amit K~Roy
  Chowdhury}, {and} \bibinfo{person}{Ananthram Swami}.}
  \bibinfo{year}{2018}\natexlab{}.
\newblock \showarticletitle{Adversarial perturbations against real-time video
  classification systems}.
\newblock \bibinfo{journal}{\emph{arXiv preprint arXiv:1807.00458}}
  (\bibinfo{year}{2018}).
\newblock


\bibitem[\protect\citeauthoryear{Liu, Wen, Yu, Li, Raj, and Song}{Liu
  et~al\mbox{.}}{2017}]%
        {liu2017sphereface}
\bibfield{author}{\bibinfo{person}{Weiyang Liu}, \bibinfo{person}{Yandong Wen},
  \bibinfo{person}{Zhiding Yu}, \bibinfo{person}{Ming Li},
  \bibinfo{person}{Bhiksha Raj}, {and} \bibinfo{person}{Le Song}.}
  \bibinfo{year}{2017}\natexlab{}.
\newblock \showarticletitle{Sphereface: Deep hypersphere embedding for face
  recognition}. In \bibinfo{booktitle}{\emph{Proceedings of the IEEE conference
  on computer vision and pattern recognition}}. \bibinfo{pages}{212--220}.
\newblock


\bibitem[\protect\citeauthoryear{Liu, Chen, Liu, and Song}{Liu
  et~al\mbox{.}}{2016}]%
        {liu2016delving}
\bibfield{author}{\bibinfo{person}{Yanpei Liu}, \bibinfo{person}{Xinyun Chen},
  \bibinfo{person}{Chang Liu}, {and} \bibinfo{person}{Dawn Song}.}
  \bibinfo{year}{2016}\natexlab{}.
\newblock \showarticletitle{Delving into transferable adversarial examples and
  black-box attacks}.
\newblock \bibinfo{journal}{\emph{arXiv preprint arXiv:1611.02770}}
  (\bibinfo{year}{2016}).
\newblock


\bibitem[\protect\citeauthoryear{Lo and Patel}{Lo and Patel}{2020}]%
        {lo2020defending}
\bibfield{author}{\bibinfo{person}{Shao-Yuan Lo} {and}
  \bibinfo{person}{Vishal~M Patel}.} \bibinfo{year}{2020}\natexlab{}.
\newblock \showarticletitle{Defending against multiple and unforeseen
  adversarial videos}.
\newblock \bibinfo{journal}{\emph{arXiv preprint arXiv:2009.05244}}
  (\bibinfo{year}{2020}).
\newblock


\bibitem[\protect\citeauthoryear{Lu, Issaranon, and Forsyth}{Lu
  et~al\mbox{.}}{2017}]%
        {lu2017safetynet}
\bibfield{author}{\bibinfo{person}{Jiajun Lu}, \bibinfo{person}{Theerasit
  Issaranon}, {and} \bibinfo{person}{David Forsyth}.}
  \bibinfo{year}{2017}\natexlab{}.
\newblock \showarticletitle{Safetynet: Detecting and rejecting adversarial
  examples robustly}. In \bibinfo{booktitle}{\emph{Proceedings of the IEEE
  International Conference on Computer Vision}}. \bibinfo{pages}{446--454}.
\newblock


\bibitem[\protect\citeauthoryear{Madry, Makelov, Schmidt, Tsipras, and
  Vladu}{Madry et~al\mbox{.}}{2017}]%
        {madry2017towards}
\bibfield{author}{\bibinfo{person}{Aleksander Madry},
  \bibinfo{person}{Aleksandar Makelov}, \bibinfo{person}{Ludwig Schmidt},
  \bibinfo{person}{Dimitris Tsipras}, {and} \bibinfo{person}{Adrian Vladu}.}
  \bibinfo{year}{2017}\natexlab{}.
\newblock \showarticletitle{Towards deep learning models resistant to
  adversarial attacks}.
\newblock \bibinfo{journal}{\emph{arXiv preprint arXiv:1706.06083}}
  (\bibinfo{year}{2017}).
\newblock


\bibitem[\protect\citeauthoryear{Materzynska, Berger, Bax, and
  Memisevic}{Materzynska et~al\mbox{.}}{2019}]%
        {materzynska2019jester}
\bibfield{author}{\bibinfo{person}{Joanna Materzynska},
  \bibinfo{person}{Guillaume Berger}, \bibinfo{person}{Ingo Bax}, {and}
  \bibinfo{person}{Roland Memisevic}.} \bibinfo{year}{2019}\natexlab{}.
\newblock \showarticletitle{The jester dataset: A large-scale video dataset of
  human gestures}. In \bibinfo{booktitle}{\emph{Proc. IEEE Intl. Conf. on
  Computer Vision Workshops}}. \bibinfo{pages}{0--0}.
\newblock


\bibitem[\protect\citeauthoryear{Meng and Chen}{Meng and Chen}{2017}]%
        {meng2017magnet}
\bibfield{author}{\bibinfo{person}{Dongyu Meng} {and} \bibinfo{person}{Hao
  Chen}.} \bibinfo{year}{2017}\natexlab{}.
\newblock \showarticletitle{Magnet: a two-pronged defense against adversarial
  examples}. In \bibinfo{booktitle}{\emph{Proceedings of the 2017 ACM SIGSAC
  conference on computer and communications security}}.
  \bibinfo{pages}{135--147}.
\newblock


\bibitem[\protect\citeauthoryear{Moosavi-Dezfooli, Fawzi, Fawzi, and
  Frossard}{Moosavi-Dezfooli et~al\mbox{.}}{2017}]%
        {moosavi2017universal}
\bibfield{author}{\bibinfo{person}{Seyed-Mohsen Moosavi-Dezfooli},
  \bibinfo{person}{Alhussein Fawzi}, \bibinfo{person}{Omar Fawzi}, {and}
  \bibinfo{person}{Pascal Frossard}.} \bibinfo{year}{2017}\natexlab{}.
\newblock \showarticletitle{Universal adversarial perturbations}. In
  \bibinfo{booktitle}{\emph{Proceedings of the IEEE conference on computer
  vision and pattern recognition}}. \bibinfo{pages}{1765--1773}.
\newblock


\bibitem[\protect\citeauthoryear{Moosavi-Dezfooli, Fawzi, and
  Frossard}{Moosavi-Dezfooli et~al\mbox{.}}{2016}]%
        {moosavi2016deepfool}
\bibfield{author}{\bibinfo{person}{Seyed-Mohsen Moosavi-Dezfooli},
  \bibinfo{person}{Alhussein Fawzi}, {and} \bibinfo{person}{Pascal Frossard}.}
  \bibinfo{year}{2016}\natexlab{}.
\newblock \showarticletitle{Deepfool: a simple and accurate method to fool deep
  neural networks}. In \bibinfo{booktitle}{\emph{Proceedings of the IEEE
  conference on computer vision and pattern recognition}}.
  \bibinfo{pages}{2574--2582}.
\newblock


\bibitem[\protect\citeauthoryear{Newell, Yang, and Deng}{Newell
  et~al\mbox{.}}{2016}]%
        {newell2016stacked}
\bibfield{author}{\bibinfo{person}{Alejandro Newell}, \bibinfo{person}{Kaiyu
  Yang}, {and} \bibinfo{person}{Jia Deng}.} \bibinfo{year}{2016}\natexlab{}.
\newblock \showarticletitle{Stacked hourglass networks for human pose
  estimation}. In \bibinfo{booktitle}{\emph{European conference on computer
  vision}}. Springer, \bibinfo{pages}{483--499}.
\newblock


\bibitem[\protect\citeauthoryear{Ondruska and Posner}{Ondruska and
  Posner}{2016}]%
        {ondruska2016deep}
\bibfield{author}{\bibinfo{person}{Peter Ondruska} {and}
  \bibinfo{person}{Ingmar Posner}.} \bibinfo{year}{2016}\natexlab{}.
\newblock \showarticletitle{Deep tracking: Seeing beyond seeing using recurrent
  neural networks}. In \bibinfo{booktitle}{\emph{Thirtieth AAAI Conference on
  Artificial Intelligence}}.
\newblock


\bibitem[\protect\citeauthoryear{Papernot, McDaniel, and Goodfellow}{Papernot
  et~al\mbox{.}}{2016a}]%
        {papernot2016transferability}
\bibfield{author}{\bibinfo{person}{Nicolas Papernot}, \bibinfo{person}{Patrick
  McDaniel}, {and} \bibinfo{person}{Ian Goodfellow}.}
  \bibinfo{year}{2016}\natexlab{a}.
\newblock \showarticletitle{Transferability in machine learning: from phenomena
  to black-box attacks using adversarial samples}.
\newblock \bibinfo{journal}{\emph{arXiv preprint arXiv:1605.07277}}
  (\bibinfo{year}{2016}).
\newblock


\bibitem[\protect\citeauthoryear{Papernot, McDaniel, Goodfellow, Jha, Celik,
  and Swami}{Papernot et~al\mbox{.}}{2017}]%
        {papernot2017practical}
\bibfield{author}{\bibinfo{person}{Nicolas Papernot}, \bibinfo{person}{Patrick
  McDaniel}, \bibinfo{person}{Ian Goodfellow}, \bibinfo{person}{Somesh Jha},
  \bibinfo{person}{Z~Berkay Celik}, {and} \bibinfo{person}{Ananthram Swami}.}
  \bibinfo{year}{2017}\natexlab{}.
\newblock \showarticletitle{Practical black-box attacks against machine
  learning}. In \bibinfo{booktitle}{\emph{Proceedings of the 2017 ACM on Asia
  conference on computer and communications security}}.
  \bibinfo{pages}{506--519}.
\newblock


\bibitem[\protect\citeauthoryear{Papernot, McDaniel, Jha, Fredrikson, Celik,
  and Swami}{Papernot et~al\mbox{.}}{2016b}]%
        {papernot2016limitations}
\bibfield{author}{\bibinfo{person}{Nicolas Papernot}, \bibinfo{person}{Patrick
  McDaniel}, \bibinfo{person}{Somesh Jha}, \bibinfo{person}{Matt Fredrikson},
  \bibinfo{person}{Z~Berkay Celik}, {and} \bibinfo{person}{Ananthram Swami}.}
  \bibinfo{year}{2016}\natexlab{b}.
\newblock \showarticletitle{The limitations of deep learning in adversarial
  settings}. In \bibinfo{booktitle}{\emph{2016 IEEE European symposium on
  security and privacy (EuroS\&P)}}. IEEE, \bibinfo{pages}{372--387}.
\newblock


\bibitem[\protect\citeauthoryear{Papernot, McDaniel, Wu, Jha, and
  Swami}{Papernot et~al\mbox{.}}{2016c}]%
        {papernot2016distillation}
\bibfield{author}{\bibinfo{person}{Nicolas Papernot}, \bibinfo{person}{Patrick
  McDaniel}, \bibinfo{person}{Xi Wu}, \bibinfo{person}{Somesh Jha}, {and}
  \bibinfo{person}{Ananthram Swami}.} \bibinfo{year}{2016}\natexlab{c}.
\newblock \showarticletitle{Distillation as a defense to adversarial
  perturbations against deep neural networks}. In
  \bibinfo{booktitle}{\emph{2016 IEEE Symposium on Security and Privacy (SP)}}.
  IEEE, \bibinfo{pages}{582--597}.
\newblock


\bibitem[\protect\citeauthoryear{Rey-de Castro and Rabitz}{Rey-de Castro and
  Rabitz}{2018}]%
        {rey2018targeted}
\bibfield{author}{\bibinfo{person}{Roberto Rey-de Castro} {and}
  \bibinfo{person}{Herschel Rabitz}.} \bibinfo{year}{2018}\natexlab{}.
\newblock \showarticletitle{Targeted nonlinear adversarial perturbations in
  images and videos}.
\newblock \bibinfo{journal}{\emph{arXiv preprint arXiv:1809.00958}}
  (\bibinfo{year}{2018}).
\newblock


\bibitem[\protect\citeauthoryear{Samangouei, Kabkab, and Chellappa}{Samangouei
  et~al\mbox{.}}{2018}]%
        {samangouei2018defense}
\bibfield{author}{\bibinfo{person}{Pouya Samangouei}, \bibinfo{person}{Maya
  Kabkab}, {and} \bibinfo{person}{Rama Chellappa}.}
  \bibinfo{year}{2018}\natexlab{}.
\newblock \showarticletitle{Defense-gan: Protecting classifiers against
  adversarial attacks using generative models}.
\newblock \bibinfo{journal}{\emph{arXiv preprint arXiv:1805.06605}}
  (\bibinfo{year}{2018}).
\newblock


\bibitem[\protect\citeauthoryear{Schroff, Kalenichenko, and Philbin}{Schroff
  et~al\mbox{.}}{2015}]%
        {schroff2015facenet}
\bibfield{author}{\bibinfo{person}{Florian Schroff}, \bibinfo{person}{Dmitry
  Kalenichenko}, {and} \bibinfo{person}{James Philbin}.}
  \bibinfo{year}{2015}\natexlab{}.
\newblock \showarticletitle{Facenet: A unified embedding for face recognition
  and clustering}. In \bibinfo{booktitle}{\emph{Proc. IEEE Conf. Computer
  Vision and Pattern Recognition}}. \bibinfo{pages}{815--823}.
\newblock


\bibitem[\protect\citeauthoryear{Simonyan and Zisserman}{Simonyan and
  Zisserman}{2014}]%
        {simonyan2014two}
\bibfield{author}{\bibinfo{person}{Karen Simonyan} {and}
  \bibinfo{person}{Andrew Zisserman}.} \bibinfo{year}{2014}\natexlab{}.
\newblock \showarticletitle{Two-stream convolutional networks for action
  recognition in videos}. In \bibinfo{booktitle}{\emph{Advances in neural
  information processing systems}}. \bibinfo{pages}{568--576}.
\newblock


\bibitem[\protect\citeauthoryear{Song, Kim, Nowozin, Ermon, and Kushman}{Song
  et~al\mbox{.}}{2017}]%
        {song2017pixeldefend}
\bibfield{author}{\bibinfo{person}{Yang Song}, \bibinfo{person}{Taesup Kim},
  \bibinfo{person}{Sebastian Nowozin}, \bibinfo{person}{Stefano Ermon}, {and}
  \bibinfo{person}{Nate Kushman}.} \bibinfo{year}{2017}\natexlab{}.
\newblock \showarticletitle{Pixeldefend: Leveraging generative models to
  understand and defend against adversarial examples}.
\newblock \bibinfo{journal}{\emph{arXiv preprint arXiv:1710.10766}}
  (\bibinfo{year}{2017}).
\newblock


\bibitem[\protect\citeauthoryear{Soomro, Zamir, and Shah}{Soomro
  et~al\mbox{.}}{2012}]%
        {soomro2012ucf101}
\bibfield{author}{\bibinfo{person}{Khurram Soomro},
  \bibinfo{person}{Amir~Roshan Zamir}, {and} \bibinfo{person}{Mubarak Shah}.}
  \bibinfo{year}{2012}\natexlab{}.
\newblock \showarticletitle{UCF101: A dataset of 101 human actions classes from
  videos in the wild}.
\newblock \bibinfo{journal}{\emph{arXiv preprint arXiv:1212.0402}}
  (\bibinfo{year}{2012}).
\newblock


\bibitem[\protect\citeauthoryear{Su, Vargas, and Sakurai}{Su
  et~al\mbox{.}}{2019}]%
        {su2019one}
\bibfield{author}{\bibinfo{person}{Jiawei Su},
  \bibinfo{person}{Danilo~Vasconcellos Vargas}, {and} \bibinfo{person}{Kouichi
  Sakurai}.} \bibinfo{year}{2019}\natexlab{}.
\newblock \showarticletitle{One pixel attack for fooling deep neural networks}.
\newblock \bibinfo{journal}{\emph{IEEE Transactions on Evolutionary
  Computation}} \bibinfo{volume}{23}, \bibinfo{number}{5}
  (\bibinfo{year}{2019}), \bibinfo{pages}{828--841}.
\newblock


\bibitem[\protect\citeauthoryear{Sudhakaran and Lanz}{Sudhakaran and
  Lanz}{2017}]%
        {sudhakaran2017learning}
\bibfield{author}{\bibinfo{person}{Swathikiran Sudhakaran} {and}
  \bibinfo{person}{Oswald Lanz}.} \bibinfo{year}{2017}\natexlab{}.
\newblock \showarticletitle{Learning to detect violent videos using
  convolutional long short-term memory}. In \bibinfo{booktitle}{\emph{2017 14th
  IEEE International Conference on Advanced Video and Signal Based Surveillance
  (AVSS)}}. IEEE, \bibinfo{pages}{1--6}.
\newblock


\bibitem[\protect\citeauthoryear{Szegedy, Zaremba, Sutskever, Bruna, Erhan,
  Goodfellow, and Fergus}{Szegedy et~al\mbox{.}}{2013}]%
        {szegedy2013intriguing}
\bibfield{author}{\bibinfo{person}{Christian Szegedy},
  \bibinfo{person}{Wojciech Zaremba}, \bibinfo{person}{Ilya Sutskever},
  \bibinfo{person}{Joan Bruna}, \bibinfo{person}{Dumitru Erhan},
  \bibinfo{person}{Ian Goodfellow}, {and} \bibinfo{person}{Rob Fergus}.}
  \bibinfo{year}{2013}\natexlab{}.
\newblock \showarticletitle{Intriguing properties of neural networks}.
\newblock \bibinfo{journal}{\emph{arXiv preprint arXiv:1312.6199}}
  (\bibinfo{year}{2013}).
\newblock


\bibitem[\protect\citeauthoryear{Tram{\`e}r, Kurakin, Papernot, Goodfellow,
  Boneh, and McDaniel}{Tram{\`e}r et~al\mbox{.}}{2017}]%
        {tramer2017ensemble}
\bibfield{author}{\bibinfo{person}{Florian Tram{\`e}r}, \bibinfo{person}{Alexey
  Kurakin}, \bibinfo{person}{Nicolas Papernot}, \bibinfo{person}{Ian
  Goodfellow}, \bibinfo{person}{Dan Boneh}, {and} \bibinfo{person}{Patrick
  McDaniel}.} \bibinfo{year}{2017}\natexlab{}.
\newblock \showarticletitle{Ensemble adversarial training: Attacks and
  defenses}.
\newblock \bibinfo{journal}{\emph{arXiv preprint arXiv:1705.07204}}
  (\bibinfo{year}{2017}).
\newblock


\bibitem[\protect\citeauthoryear{Wang, Jabri, and Efros}{Wang
  et~al\mbox{.}}{2019a}]%
        {wang2019learning}
\bibfield{author}{\bibinfo{person}{Xiaolong Wang}, \bibinfo{person}{Allan
  Jabri}, {and} \bibinfo{person}{Alexei~A Efros}.}
  \bibinfo{year}{2019}\natexlab{a}.
\newblock \showarticletitle{Learning correspondence from the cycle-consistency
  of time}. In \bibinfo{booktitle}{\emph{Proceedings of the IEEE Conference on
  Computer Vision and Pattern Recognition}}. \bibinfo{pages}{2566--2576}.
\newblock


\bibitem[\protect\citeauthoryear{Wang, Xu, Liu, Zhu, and Shao}{Wang
  et~al\mbox{.}}{2019b}]%
        {wang2019ranet}
\bibfield{author}{\bibinfo{person}{Ziqin Wang}, \bibinfo{person}{Jun Xu},
  \bibinfo{person}{Li Liu}, \bibinfo{person}{Fan Zhu}, {and}
  \bibinfo{person}{Ling Shao}.} \bibinfo{year}{2019}\natexlab{b}.
\newblock \showarticletitle{Ranet: Ranking attention network for fast video
  object segmentation}. In \bibinfo{booktitle}{\emph{Proceedings of the IEEE
  international conference on computer vision}}. \bibinfo{pages}{3978--3987}.
\newblock


\bibitem[\protect\citeauthoryear{Wei, Zhu, Yuan, and Su}{Wei
  et~al\mbox{.}}{2019}]%
        {wei2019sparse}
\bibfield{author}{\bibinfo{person}{Xingxing Wei}, \bibinfo{person}{Jun Zhu},
  \bibinfo{person}{Sha Yuan}, {and} \bibinfo{person}{Hang Su}.}
  \bibinfo{year}{2019}\natexlab{}.
\newblock \showarticletitle{Sparse adversarial perturbations for videos}. In
  \bibinfo{booktitle}{\emph{Proceedings of the AAAI Conference on Artificial
  Intelligence}}, Vol.~\bibinfo{volume}{33}. \bibinfo{pages}{8973--8980}.
\newblock


\bibitem[\protect\citeauthoryear{Wei, Chen, Wei, Jiang, Chua, Zhou, and
  Jiang}{Wei et~al\mbox{.}}{2020}]%
        {wei2020heuristic}
\bibfield{author}{\bibinfo{person}{Zhipeng Wei}, \bibinfo{person}{Jingjing
  Chen}, \bibinfo{person}{Xingxing Wei}, \bibinfo{person}{Linxi Jiang},
  \bibinfo{person}{Tat-Seng Chua}, \bibinfo{person}{Fengfeng Zhou}, {and}
  \bibinfo{person}{Yu-Gang Jiang}.} \bibinfo{year}{2020}\natexlab{}.
\newblock \showarticletitle{Heuristic Black-Box Adversarial Attacks on Video
  Recognition Models.}. In \bibinfo{booktitle}{\emph{AAAI}}.
  \bibinfo{pages}{12338--12345}.
\newblock


\bibitem[\protect\citeauthoryear{Xiao, Deng, Li, Lee, Edwards, Yi, Song, Liu,
  and Molloy}{Xiao et~al\mbox{.}}{2019}]%
        {xiao2019advit}
\bibfield{author}{\bibinfo{person}{Chaowei Xiao}, \bibinfo{person}{Ruizhi
  Deng}, \bibinfo{person}{Bo Li}, \bibinfo{person}{Taesung Lee},
  \bibinfo{person}{Benjamin Edwards}, \bibinfo{person}{Jinfeng Yi},
  \bibinfo{person}{Dawn Song}, \bibinfo{person}{Mingyan Liu}, {and}
  \bibinfo{person}{Ian Molloy}.} \bibinfo{year}{2019}\natexlab{}.
\newblock \showarticletitle{Advit: Adversarial frames identifier based on
  temporal consistency in videos}. In \bibinfo{booktitle}{\emph{Proc. IEEE
  Intl. Conf. on Computer Vision}}. \bibinfo{pages}{3968--3977}.
\newblock


\bibitem[\protect\citeauthoryear{Xie, Zhang, Zhou, Bai, Wang, Ren, and
  Yuille}{Xie et~al\mbox{.}}{2019}]%
        {xie2019improving}
\bibfield{author}{\bibinfo{person}{Cihang Xie}, \bibinfo{person}{Zhishuai
  Zhang}, \bibinfo{person}{Yuyin Zhou}, \bibinfo{person}{Song Bai},
  \bibinfo{person}{Jianyu Wang}, \bibinfo{person}{Zhou Ren}, {and}
  \bibinfo{person}{Alan~L Yuille}.} \bibinfo{year}{2019}\natexlab{}.
\newblock \showarticletitle{Improving transferability of adversarial examples
  with input diversity}. In \bibinfo{booktitle}{\emph{Proceedings of the IEEE
  Conference on Computer Vision and Pattern Recognition}}.
  \bibinfo{pages}{2730--2739}.
\newblock


\bibitem[\protect\citeauthoryear{Zeng, Liu, Wang, Qiu, Xie, Tai, Tang, and
  Yuille}{Zeng et~al\mbox{.}}{2019}]%
        {zeng2019adversarial}
\bibfield{author}{\bibinfo{person}{Xiaohui Zeng}, \bibinfo{person}{Chenxi Liu},
  \bibinfo{person}{Yu-Siang Wang}, \bibinfo{person}{Weichao Qiu},
  \bibinfo{person}{Lingxi Xie}, \bibinfo{person}{Yu-Wing Tai},
  \bibinfo{person}{Chi-Keung Tang}, {and} \bibinfo{person}{Alan~L Yuille}.}
  \bibinfo{year}{2019}\natexlab{}.
\newblock \showarticletitle{Adversarial attacks beyond the image space}. In
  \bibinfo{booktitle}{\emph{Proceedings of the IEEE Conference on Computer
  Vision and Pattern Recognition}}. \bibinfo{pages}{4302--4311}.
\newblock


\bibitem[\protect\citeauthoryear{Zhang, Fidler, and Urtasun}{Zhang
  et~al\mbox{.}}{2016}]%
        {zhang2016instance}
\bibfield{author}{\bibinfo{person}{Ziyu Zhang}, \bibinfo{person}{Sanja Fidler},
  {and} \bibinfo{person}{Raquel Urtasun}.} \bibinfo{year}{2016}\natexlab{}.
\newblock \showarticletitle{Instance-level segmentation for autonomous driving
  with deep densely connected mrfs}. In \bibinfo{booktitle}{\emph{Proceedings
  of the IEEE Conference on Computer Vision and Pattern Recognition}}.
  \bibinfo{pages}{669--677}.
\newblock


\bibitem[\protect\citeauthoryear{Zhao, Dua, and Singh}{Zhao
  et~al\mbox{.}}{2017}]%
        {zhao2017generating}
\bibfield{author}{\bibinfo{person}{Zhengli Zhao}, \bibinfo{person}{Dheeru Dua},
  {and} \bibinfo{person}{Sameer Singh}.} \bibinfo{year}{2017}\natexlab{}.
\newblock \showarticletitle{Generating natural adversarial examples}.
\newblock \bibinfo{journal}{\emph{arXiv preprint arXiv:1710.11342}}
  (\bibinfo{year}{2017}).
\newblock


\end{thebibliography}

\end{document}